%% file: main.tex
\documentclass[conference]{IEEEtran}
\IEEEoverridecommandlockouts

\usepackage{booktabs,tabularx}
\usepackage{tikz}
\usetikzlibrary{arrows.meta,positioning,calc,decorations.pathreplacing}
\usepackage[colorlinks=true, linkcolor=black, citecolor=black, urlcolor=black]{hyperref}

\usepackage{algorithm}
\usepackage[noend]{algpseudocode} 
\makeatletter
\newcommand\fs@betterruled{%
  \def\@fs@cfont{\small}%
  \let\@fs@capt\floatc@ruled
  \def\@fs@pre{\vspace*{5pt}\hrule height .8pt depth 0pt \kern2pt}%
  \def\@fs@post{\kern2pt\hrule\relax\vspace*{-1em}}%
  \def\@fs@mid{\kern2pt\hrule\kern2pt}%
  \let\@fs@iftopcapt\iftrue
}
\floatstyle{betterruled}
\restylefloat{algorithm}
\renewcommand{\fnum@algorithm}{\small\bfseries Algorithm~\thealgorithm}
\renewcommand{\Comment}[1]{\hfill{\footnotesize$\triangleright$~#1}}
\algrenewcommand\algorithmicrequire{\textbf{Input:}}
\makeatother

\usepackage{titlesec}
\usepackage{indentfirst} 
\titlespacing{\subsubsection}{\parindent}{*1}{0pt}
\usepackage{booktabs}
\usepackage{colortbl,xcolor}
\usepackage{dblfloatfix}

\usepackage{enumitem}

\usepackage{cite}
\usepackage{amsmath,amssymb,amsfonts}
\usepackage{graphicx}
\usepackage{textcomp}
\usepackage{xcolor}

\graphicspath{{images/}}

\def\BibTeX{{\rm B\kern-.05em{\sc i\kern-.025em b}\kern-.08em
    T\kern-.1667em\lower.7ex\hbox{E}\kern-.125emX}}

\begin{document}

\pdfpagewidth=8.5in
\pdfpageheight=11in



\pagenumbering{arabic}

\title{Snowball: A Scalable All-to-All Ising Machine with Dual-Mode Markov Chain Monte Carlo Spin Selection and Asynchronous Spin Updates for Fast Combinatorial Optimization}

\author{
\IEEEauthorblockN{Seungki Hong \qquad \qquad Kyeongwon Jeong \qquad \qquad Taekwang Jang}
\IEEEauthorblockA{ETH Zurich}\vspace{-2em}
}


\maketitle
\thispagestyle{plain}
\pagestyle{plain}

\bstctlcite{IEEEexample:BSTcontrol} 

\input{sections/abstract}
\input{sections/introduction}
\input{sections/background}
\input{sections/motivation}
\input{sections/snowball}
\input{sections/evaluation}
\input{sections/conclusion}


\bibliographystyle{IEEEtranS}
\bibliography{refs}

\end{document}

%% file: sections/abstract.tex
\begin{abstract}
Ising machines have emerged as accelerators for combinatorial optimization. To enable practical deployment, this work aims to reduce time-to-solution by addressing three challenges: (1) hardware topology, (2) spin selection and update algorithms, and (3) scalable coupling-coefficient precision. Restricted topologies require minor embedding; naive parallel updates can oscillate or stall; and limited precision can preclude feasible mappings or degrade solution quality.

This work presents {\em Snowball}, a digital, scalable, all-to-all coupled Ising machine that integrates dual-mode Markov chain Monte Carlo spin selection with asynchronous spin updates to promote convergence and reduce time-to-solution. The digital architecture supports wide, configurable coupling precision, unlike many analog realizations at high bit widths. A prototype on an AMD Alveo U250 accelerator card achieves an 8$\times$ reduction in time-to-solution relative to a state-of-the-art Ising machine on the same benchmark instance.
\end{abstract}

%% file: sections/introduction.tex
\section{Introduction}
Many hard combinatorial optimization problems can be mapped to the Ising model and approximately solved using specialized hardware known as Ising machines~\cite{doi:10.1126/science.aah4243, goto2019combinatorial, 9162869, boothby2020next, 9062965, 10609617, 10067504, 10454294, 10454272, 10185207, 11074989, 10719421, 9980631, 10849084, 9062938, 9431401, 9508725, 10849161, 10121286, 10121303, 9731680, 10067622, 10454340, 11075071, 7424772, 8776512, 8780296, 9365748, 9634769, 9394375}. Such machines have attracted significant attention due to their potential to deliver low time-to-solution. However, realizing scalable Ising machines with all-to-all connectivity, well-behaved convergence of Ising dynamics, and high-resolution, scalable coupling coefficients remains challenging, particularly for large, densely connected problem instances.

This work introduces Snowball, a digital, scalable, all-to-all coupled Ising machine that integrates dual-mode Markov chain Monte Carlo spin selection and asynchronous spin updates and is implemented on an AMD Alveo U250 accelerator card. The architecture is designed to support efficient exploration and exploitation, reduce embedding overhead through effective all-to-all coupling, and exploit the parallelism available in modern FPGA-based accelerators. Experimental results demonstrate that Snowball can substantially reduce time-to-solution compared to state-of-the-art Ising machines while maintaining competitive solution quality.

The remainder of this paper is organized as follows. Section~\ref{sec:background} reviews the basic foundations of combinatorial optimization and the Ising model, with an emphasis on Ising dynamics. Section~\ref{sec:motivation} analyzes key design considerations related to spin-connectivity topology, the convergence behavior of Ising dynamics, and the precision of coupling coefficients, thereby motivating the need for a digital, all-to-all coupled architecture with dual-mode operation. Section~\ref{sec:snowball} details the Snowball architecture, including the dual-mode Markov chain Monte Carlo spin-selection mechanism, the asynchronous update pipeline, and its realization on the AMD Alveo U250 accelerator card.
Section~\ref{sec:evaluation} presents an experimental evaluation on representative combinatorial optimization benchmarks, demonstrating the solution quality and time-to-solution benefits of Snowball over state-of-the-art Ising machines. This paper makes the following key contributions:
\begin{itemize}[leftmargin=*,topsep=0pt,itemsep=0pt]
\item Systematic analysis for Ising machines. The paper identifies three key design considerations for Ising machines for combinatorial optimization: (1) all-to-all spin connectivity, which is critical for low time-to-solution due to reduced embedding overhead and richer spin interactions; (2) carefully designed single-spin updates, which can converge faster than naive parallel-spin updates; and (3) sufficient coupling-coefficient precision for robust performance across diverse problem instances.

\item Efficient and scalable Snowball architecture. Snowball employs bit-plane decomposition for scalable coupling coefficients and row-/column-major on-chip buffering with incremental updates to reduce memory footprint and traffic, thereby lowering time-to-solution on an AMD Alveo U250 accelerator.

\item On-chip dual-mode Markov chain Monte Carlo. A dual-mode Markov chain Monte Carlo spin-selection mechanism is implemented on chip, leveraging stateless random number generation and a piecewise-linear approximation of the exponential function to implement Ising dynamics efficiently in hardware, without prohibitive area or latency overhead.

\item Evaluation on real hardware. Snowball achieves an $8\times$ reduction in time-to-solution compared to a state-of-the-art Ising machine in real-hardware experiments.
\end{itemize}

%% file: sections/background.tex
\section{Background}\label{sec:background}
\subsection{Combinatorial Optimization}
Combinatorial optimization concerns the minimization or maximization of an objective over a discrete (typically finite) feasible set. A combinatorial optimization problem consists of concrete input data that determine a feasible set \(X\) and an objective function \(f\); the task is to compute \(\max\{\,f(x): x \in X\,\}\), with the analogous definition for minimization. Many canonical problems are NP-hard. Representative NP-hard problems include Max-Cut and graph partitioning. 
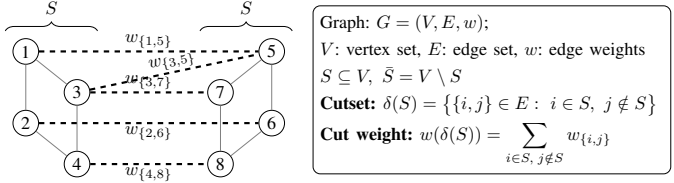
\begin{figure}[h]
\centering
\vspace{-0.7em}
\resizebox{\columnwidth}{!}{%
\begin{tikzpicture}[
  vertex/.style={circle,draw,inner sep=0pt,minimum size=5mm,fill=white,},
  intra/.style={gray},
  cutedge/.style={very thick,dashed},
  note/.style={rounded corners,draw,fill=white,inner sep=4pt,align=left}
]

\node[vertex] (a) at (-2.4,  1.2) {1};
\node[vertex] (b) at (-2.4, -0.2) {2};
\node[vertex] (c) at (-1.4,  0.4) {3};
\node[vertex] (d) at (-1.4, -1.0) {4};

\node[vertex] (e) at ( 2.4,  1.2) {5};
\node[vertex] (f) at ( 2.4, -0.2) {6};
\node[vertex] (g) at ( 1.4,  0.4) {7};
\node[vertex] (h) at ( 1.4, -1.0) {8};

\foreach \u/\v in {a/b,a/c,b/d,c/d,e/f,e/g,f/h,g/h}
  \draw[intra] (\u)--(\v);

\draw[cutedge] (a)-- node[above,sloped,inner sep=1pt] {$w_{\{1,5\}}$} (e);
\draw[cutedge] (c)-- node[above,sloped,inner sep=1pt] {$w_{\{3,5\}}$} (e);
\draw[cutedge] (b)-- node[below,sloped,inner sep=1pt] {$w_{\{2,6\}}$} (f);
\draw[cutedge] (c)-- node[above,sloped,inner sep=1pt] {$w_{\{3,7\}}$} (g);
\draw[cutedge] (d)-- node[below,sloped,inner sep=1pt] {$w_{\{4,8\}}$} (h);


\draw[decorate,decoration={brace,amplitude=5pt}] (-2.8,1.6) -- (-1.0,1.6)
  node[midway,above=6pt] {$S$};
\draw[decorate,decoration={brace,amplitude=5pt}] (1.0,1.6) -- (2.8,1.6)
  node[midway,above=6pt] {$\bar S$};

\node[note,anchor=west] at (3.2,0.45) {%
  \(\begin{aligned}
  &\text{Graph: }G=(V,E,w)\text{;}\\[0pt]
  &V\text{: vertex set, }E\text{: edge set, }w\text{: edge weights}\\[0pt]
  &S\subseteq V,\ \bar S=V\setminus S\\[0pt]
  &\text{\textbf{Cutset: }}\delta(S)=\bigl\{\{i,j\}\in E:\ i\in S,\ j\notin S\bigr\}\\[0pt]
  &\text{\textbf{Cut weight: }} w(\delta(S))=\sum_{i\in S,\; j\notin S} w_{\{i,j\}}\\[0pt]
  \end{aligned}\)
};

\end{tikzpicture}%
}
\vspace{-2em}
\caption{Illustration of a cut induced by \(S\subseteq V\). Dashed edges form the cutset \(\delta(S)\); their weights sum to the cut weight \(w(\delta(S))\).}
\label{fig:maxcut}
\vspace{-0.3em}
\end{figure}

The Max-Cut problem seeks a partition of the vertex set \(V\) into two disjoint sets \(S\) and \(\bar S\) that maximizes the total weight of edges crossing between them. Figure~\ref{fig:maxcut} shows the bipartition \((S,\bar S)\) and the corresponding cutset \(\delta(S)\); only edges with one endpoint in \(S\) and the other in \(\bar S\) contribute to the cut weight \(w(\delta(S))\), where each edge \(\{i,j\}\in E\) has a nonnegative weight \(w_{\{i,j\}}\). Applications include clustering/community detection on signed graphs~\cite{bansal2004correlation, kunegis2010spectral}, and spin-glass ground-state estimation~\cite{barahona1982computational}.
The graph partitioning problem seeks a balanced partition of \(V\) into two disjoint sets \(S\) and \(\bar S\) that minimizes the total weight of edges crossing between them. Each edge \(\{i,j\}\in E\) has a nonnegative weight \(w_{\{i,j\}}\). Applications include load balancing and communication minimization in parallel scientific computing~\cite{hendrickson2000graph}, image segmentation~\cite{boykov2001interactive}, and VLSI placement~\cite{dunlop2004procedure}.


\subsection{Ising Machine}
The Ising model is a mathematical model of interacting binary variables, called spins. Each spin $s_i$ takes one of two values, $s_i \in \{-1,+1\}$, where $-1$ denotes spin down and $+1$ denotes spin up. The spins are placed on the vertices of a graph $G=(V,E)$. Pairs of spins that are connected by an edge $\{i,j\}\in E$ interact with each other. Each interacting pair $\{i,j\}$ is associated with a coupling $J_{ij}$, and each spin $i$ is associated with an external field $h_i$. For a spin configuration $\mathbf{s} = (s_1,\ldots,s_n)$, the Ising Hamiltonian (energy function) is defined as
\begin{equation}
H(\mathbf{s}) = -\sum_{i<j} J_{ij}\, s_i s_j - \sum_i h_i s_i .
\label{eq:ising}
\end{equation}
The ground state $\mathbf{s}^\star$ is the configuration with the lowest energy, $\mathbf{s}^\star = \arg\min_{\mathbf{s}\in\{-1,+1\}^n} H(\mathbf{s})$.

\begin{figure}[h]
  \centering
  \includegraphics[width=\columnwidth]{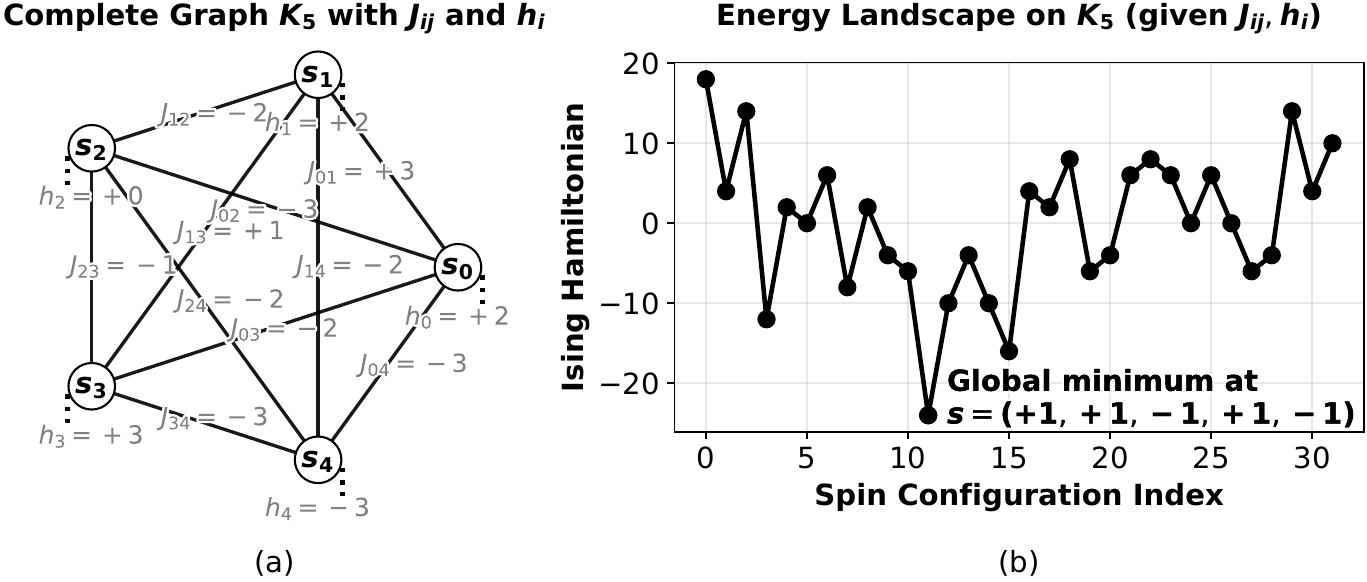}
  \vspace{-2em}
  \caption{(a) Fully connected five-spin Ising model on the complete graph \(K_5\) with pairwise couplings \(J_{ij}\) and site-dependent external fields \(h_i\). (b) The Ising Hamiltonian is shown over all \(2^5\) spin configurations \(\mathbf{s}\in\{\pm 1\}^5\).}
  \label{fig:k5}
\end{figure}

Figure~\ref{fig:k5} illustrates a fully connected five-spin Ising model on the complete graph $K_5$ and the corresponding energy landscape over all $2^5$ spin configurations. For the parameter values used in the figure, the ground-state configuration is $\mathbf{s} = (+1,+1,-1,+1,-1)$ with energy $H(\mathbf{s}) = -\sum_{i<j} J_{ij}s_is_j - \sum_{i=1}^5 h_i s_i = -14 - 10 = -24$.

An Ising machine is a physical or algorithmic system whose intrinsic dynamics approximately minimize the Ising Hamiltonian. After the couplings $J_{ij}$ and fields $h_i$ are specified to encode a problem, the machine evolves according to its update rule and tends to settle into low-energy (ideally ground-state) spin configurations. This property is useful because many NP-hard combinatorial optimization problems (such as Max-Cut and graph partitioning) can be reformulated as the minimization of an Ising Hamiltonian. Given such a formulation, the problem instance is encoded by selecting appropriate couplings $J_{ij}$ and external fields $h_i$; the Ising machine then searches for low-energy configurations that correspond to good (or optimal) solutions.

State-of-the-art Ising machines can be categorized by physical realization: coherent optical networks based on degenerate optical parametric oscillators (e.g., CIM~\cite{doi:10.1126/science.aah4243}); FPGA implementations of simulated bifurcation that emulate adiabatic coupled-oscillator bifurcation with spin-encoded phases (e.g., SB~\cite{goto2019combinatorial}); analog CMOS coupled-oscillator arrays with spin-encoded phases that relax continuously to low-energy states~\cite{9162869}; quantum (adiabatic) annealers, which implement a time-dependent Hamiltonian whose final Hamiltonian encodes the problem Hamiltonian (e.g., D-Wave’s quantum annealer~\cite{boothby2020next}); digital annealers that perform massively parallel updates (e.g., Hitachi’s STATICA~\cite{9062965}); compute-in-memory implementations that map $J$ and $h$ directly onto memory arrays (e.g., ReAIM~\cite{10609617}); and software simulators of Ising dynamics, including multi-GPU quantum-annealing simulators~\cite{9643932} and simulated annealing (e.g., D-Wave’s Neal~\cite{dwave_neal}).

\subsection{Simulated Annealing}
\begin{figure}[h]
  \centering
  \vspace{-1.5em}
  \includegraphics[width=\columnwidth]{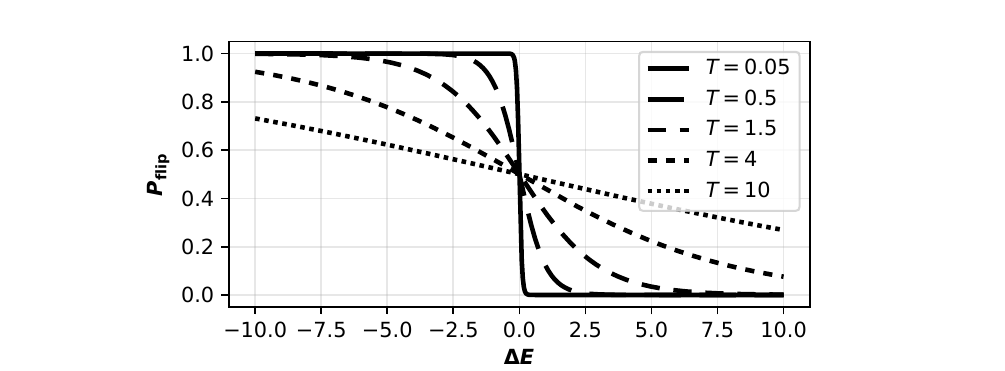}
  \vspace{-2.0em}
  \caption{Glauber flip probability $P_{\mathrm{flip}}$ versus energy change $\Delta E$. As $T\to\infty$, $P_{\mathrm{flip}}\to 0.5$ (flips are effectively random). As $T\to 0^+$, $P_{\mathrm{flip}}\to 1$ for $\Delta E<0$, $P_{\mathrm{flip}}\to 0.5$ for $\Delta E=0$, and $P_{\mathrm{flip}}\to 0$ for $\Delta E>0$—i.e., only energy-decreasing flips are accepted.}
  \label{fig:flip_p}
  \vspace{-0.5em}
\end{figure}

Simulated annealing is a probabilistic optimization method inspired by thermal annealing: by running a Markov chain at a temperature that is gradually lowered, the algorithm promotes exploration at high temperature and concentrates on low-energy states as the temperature decreases~\cite{kirkpatrick1983optimization}. To implement simulated annealing, a local single-spin update rule is required. A standard choice is Glauber dynamics~\cite{glauber1963time} or single-site heat-bath (Gibbs) updates~\cite{geman1984stochastic}, for which the flip probability is
\begin{equation}
P_{\mathrm{flip}}(s_i\to -s_i \mid \mathbf{s})=\frac{1}{1+\exp(\Delta E_i/T)} ,
\label{eq:glauber}
\end{equation} where $T$ denotes temperature, $s_i\in\{\pm1\}$, $u_i \equiv h_i+\sum_{j\neq i} J_{ij}s_j$ is the local field, and the flip energy change is defined as $\Delta E_i \equiv H(\mathbf{s}^{(i\to -i)})-H(\mathbf{s}) = 2 s_i u_i$. 
From Equation~\eqref{eq:glauber}, as $T\to\infty$ one has $P_{\mathrm{flip}}\to 0.5$ (random flips); as $T\to 0^+$, $P_{\mathrm{flip}}\to 1$ if $\Delta E_i<0$, $P_{\mathrm{flip}}\to 0.5$ if $\Delta E_i=0$, and $P_{\mathrm{flip}}\to 0$ if $\Delta E_i>0$ (only downhill moves persist). For any finite $T>0$ and any uphill move ($\Delta E_i>0$), $0<P_{\mathrm{flip}}<0.5$ (occasional uphill moves). These behaviors are illustrated in Figure~\ref{fig:flip_p}.

At any fixed $T>0$, these dynamics emulate thermal fluctuations and allow occasional uphill moves, thereby facilitating escape from local minima in rugged Ising energy landscapes.

%% file: sections/motivation.tex
\section{Motivation}\label{sec:motivation}
Motivated by pioneering CMOS- and FPGA-based Ising machine prior works from Hitachi~\cite{7063111, 7818651}, this work implements an Ising machine that, similar to Hitachi's FPGA annealing processor~\cite{7818651}, leverages simulated annealing on an AMD Zynq 7000 SoC as a proof-of-concept for a synthetic Max-Cut problem instance.

\begin{figure}[h]
  \centering
  \vspace{-0.5em}
  \includegraphics[width=\columnwidth]{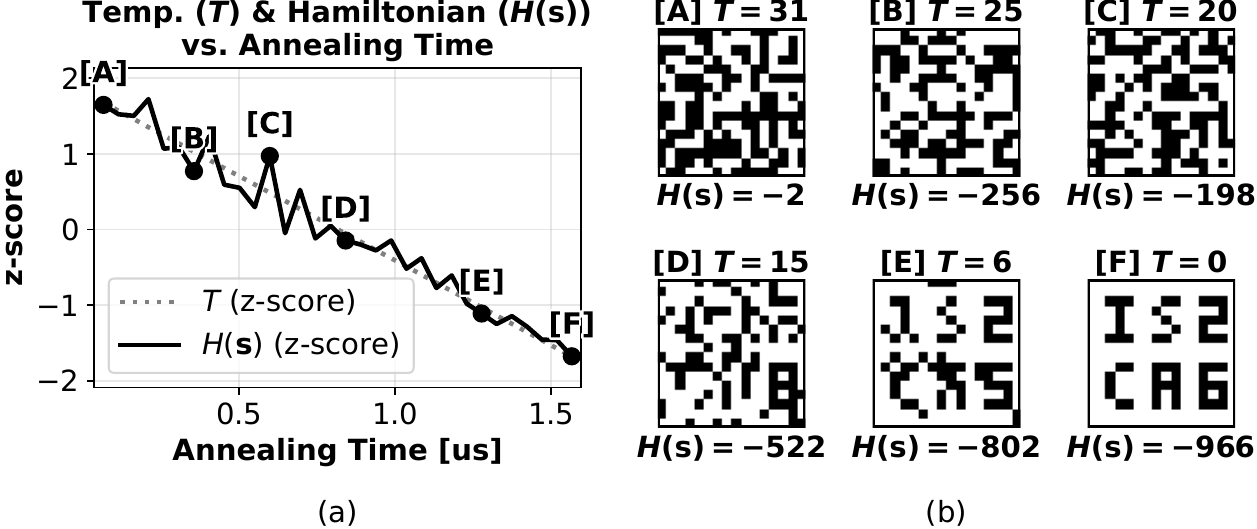}
  \vspace{-2em}
  \caption{Example run on a Max-Cut instance solved via simulated annealing. (a) Under linear cooling, \(H(\mathbf{s})\) decreases overall as \(T\) decreases. (b) Spin configurations on a 2D grid at checkpoints [A]--[F] with their \(T\) and \(H(\mathbf{s})\) values. At [F], the system attains the ground-state configuration ``ISCA26''.}
  \label{fig:bg}
  \vspace{-0.5em}
\end{figure}

Figure~\ref{fig:bg} illustrates an example run on a Max-Cut instance with a known optimum. A linear cooling schedule, in which the temperature $T$ decreases linearly with the annealing step, is applied, and the Ising Hamiltonian $H(\mathbf{s})$ tends to decrease overall as $T$ decreases. To visualize this behavior, the vertical axis shows the standardized (z-score) values of the temperature $T$ and the Ising Hamiltonian $H(\mathbf{s})$, obtained by normalizing each quantity to account for their different scales so that both can be plotted on the same axis for comparison. As shown, stochastic thermal fluctuations can cause temporary increases in energy (e.g., [B]$\to$[C]). By the end of the schedule ([F]), the system reaches the minimum energy and recovers the ground-truth string ``ISCA26''.

While prior works such as~\cite{boothby2020next, doi:10.1126/science.aah4243, 9162869, goto2019combinatorial, 10609617, 9062965, 7063111, 7818651} briefly mention certain design considerations, they mainly focus on demonstrating how their Ising machines solve combinatorial optimization problems quickly, without providing a systematic analysis. Therefore, this work systematically analyzes three key design considerations for Ising machines: (1) topology of spin connectivity, (2) convergence of Ising dynamics, and (3) precision of coupling coefficients.

\subsection{Topology of Spin Connectivity}\label{subsec:topo}
In order to solve combinatorial optimization problems such as Max-Cut, graph partitioning, and the TSP, the problem graph must be mapped onto the Ising machine. Thus, the hardware topology of the Ising machine is crucial. For example, when the problem graph is densely connected whereas the Ising machine topology is sparse, it is necessary to perform a procedure called minor embedding~\cite{choi2008minor, choi2011minor}.

\begin{figure}[h]
  \centering
  \vspace{-0.5em}
  \includegraphics[width=\columnwidth]{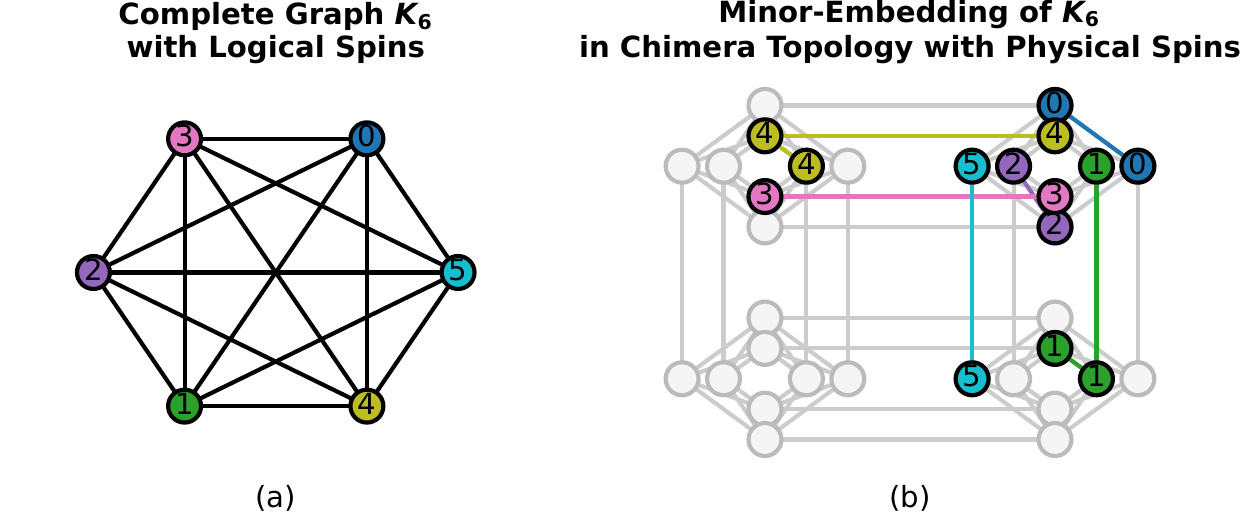}
  \vspace{-2em}
  \caption{Illustration of a minor embedding. (a) Problem graph: complete graph $K_6$ with six logical spins. (b) Corresponding embedding on the Chimera hardware topology~\cite{dwave_qpu}, requiring more than six physical spins because the hardware connectivity is sparser than that of the problem graph.}
  \label{fig:embed}
  \vspace{-0.5em}
\end{figure}

Figure~\ref{fig:embed} illustrates the minor-embedding of the complete graph $K_6$, which consists of six fully connected logical spins. In this example, a D-Wave quantum processing unit with Chimera topology~\cite{dwave_qpu} is used for illustration. Since the Chimera topology is more sparsely connected than the problem graph, additional physical spins are required. These additional physical spins are connected in chains, which are highlighted. 

The logical graph (problem graph) is a minor of the physical graph (hardware topology) if it can be obtained from the physical graph by a sequence of vertex deletions, edge deletions, and edge contractions, where contracting an edge merges its two endpoints into a single spin. Consequently, minor embedding is not always possible on a given device: even with enough physical spins, the specific connectivity and structure of the hardware topology may prevent a particular logical graph from being a minor of the hardware graph.

Therefore, an Ising machine with an all-to-all coupled spin topology is highly desirable. However, realizing all-to-all connectivity in hardware is difficult, especially as the number of spins increases. A fully connected graph of $N$ spins requires $N(N-1)/2$ couplers, i.e., $O(N^2)$ edges. This quickly leads to routing congestion on chip, which can harm signal integrity and increase crosstalk due to the dense coupling network. Thus, careful architectural and physical design is required.

Another important aspect of topology is its impact on the convergence of Ising machine algorithms. Consider two Ising machines with the same number of spins $N$, where one hardware topology is sparse and the other is dense. Even though they have the same number of spins, their convergence speed (or the number of update steps required) can differ, because a higher edge density implies that each spin interacts with more neighbors at each update. With careful algorithm design, this increased connectivity can be exploited to accelerate convergence, making an all-to-all (or denser) topology particularly attractive.

\subsection{Convergence of Ising Dynamics}\label{subsec:conv}
An Ising machine must ensure algorithmic convergence to obtain both high-quality solutions and low time-to-solution. To achieve fast convergence, the algorithm must be carefully designed. Ising machine algorithms can be divided into two main phases: spin selection and spin update. During the spin-selection phase, one or more spins are chosen, and for each selected spin the local field $u_i$ is computed (see the definition below Equation~\eqref{eq:glauber}). During the spin-update phase, one or more spins are then updated according to local update dynamics, such as Glauber dynamics, where the spin-flip probability is given by Equation~\eqref{eq:glauber} and depends on the local field.

\begin{figure}[h]
  \centering
  \vspace{-1.5em}
  \includegraphics[width=\columnwidth]{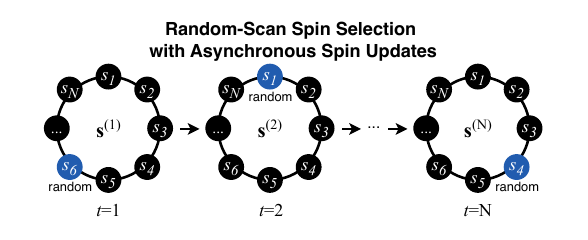}
  \vspace{-3em}
  \caption{Sequential single-spin Markov chain Monte Carlo with $N$ spins. In each iteration $t = 1,\dots,N$, one spin (highlighted in blue) is selected at random and updated according to the chosen transition rule, while the remaining spins are kept fixed, thereby generating a Markov chain over spin configurations.}
  \label{fig:smc}
  \vspace{-0.5em}
\end{figure}

In sequential Markov chain Monte Carlo, illustrated in Figure~\ref{fig:smc}, a single spin is selected uniformly at random at each iteration (random-scan spin selection) and updated sequentially along a single Markov chain (asynchronous spin updates). For the randomly selected spin, its local field is evaluated, and its state is updated according to a prescribed transition kernel (e.g., Glauber dynamics; see Equation~\eqref{eq:glauber}), which is typically designed to be ergodic and to leave a target Gibbs distribution invariant. 

On a finite state space, ergodicity means that the Markov chain is irreducible (it can reach any configuration from any initial state) and aperiodic (it does not become trapped in periodic cycles); under these conditions, the chain admits a unique stationary distribution and converges to it from any initial state, so that long-run time averages coincide with expectations under the Gibbs distribution. Since the Ising model follows a Gibbs distribution at thermal equilibrium (i.e., in the absence of net heat flow), ergodicity of the underlying Markov chain, together with invariance of the Gibbs distribution, is crucial for guaranteeing convergence of the Ising machine dynamics.

\begin{figure}[h]
  \centering
  \vspace{-0.5em}
  \includegraphics[width=\columnwidth]{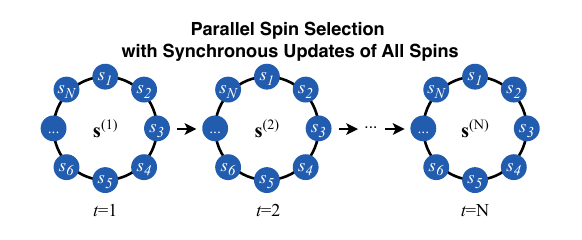}
  \vspace{-3em}
  \caption{Parallel all-spin Markov chain Monte Carlo. In each iteration, the local field $u_i$ is computed for every spin, and all spins are updated simultaneously according to a probabilistic transition rule, enabling full parallelism while still defining a Markov chain over spin configurations.}
  \label{fig:pmc}
  \vspace{-1.5em}
\end{figure}

In parallel Markov chain Monte Carlo, multiple updates are performed concurrently to exploit hardware parallelism. This can be realized either by running multiple independent Markov chains in parallel or by updating multiple spins within a single chain at the same time, as illustrated in Figure~\ref{fig:pmc}. In the latter case, the set of spins eligible for synchronous update must be chosen so as to preserve detailed balance and ergodicity.

Here, detailed balance is a condition on the transition probabilities that guarantees the target distribution is stationary and is often required when designing Markov chain Monte Carlo kernels~\cite{robert1999monte}. Let $\pi(\mathbf{s})$ denote the target Gibbs distribution over spin configurations, and let $P(\mathbf{s} \to \mathbf{s}')$ denote the transition probability from configuration $\mathbf{s}$ to $\mathbf{s}'$. The Markov chain is said to satisfy detailed balance with respect to $\pi$ if
\begin{equation}
  \pi(\mathbf{s})\, P(\mathbf{s} \to \mathbf{s}')
  = \pi(\mathbf{s}')\, P(\mathbf{s}' \to \mathbf{s})
  \quad \text{for all } \mathbf{s}, \mathbf{s}' .
  \label{eq:db}
\end{equation}
This condition implies that $\pi$ is a stationary distribution of the Markov chain, since summing the detailed-balance condition~\eqref{eq:db} over all $\mathbf{s}$ yields $\sum_{\mathbf{s}} \pi(\mathbf{s})\, P(\mathbf{s} \to \mathbf{s}') = \pi(\mathbf{s}')$ for all $\mathbf{s}'$, a condition known as global balance. Together with ergodicity, detailed balance guarantees that the chain converges to a unique stationary distribution and that time averages of observables along the chain converge to their expectations under the target Gibbs distribution.

However, when multiple spins are updated simultaneously within a single chain in parallel Markov chain Monte Carlo (Figure~\ref{fig:pmc}), detailed balance can be violated. A common strategy is to compute the local field $u_i$ for every spin from the current configuration $\mathbf{s}$ and then update all spins independently according to single-spin transition probabilities $P_i(s_i \to s_i' \mid u_i)$. This leads to a transition kernel of the form
\begin{equation}
  P(\mathbf{s} \to \mathbf{s}')
  = \prod_{i=1}^N P_i\bigl(s_i \to s_i' \mid u_i(\mathbf{s})\bigr) .
\end{equation}
For an Ising model with interacting spins, the reverse transition probability $P(\mathbf{s}' \to \mathbf{s})$ uses local fields computed from $\mathbf{s}'$, so in general
\begin{equation}
  \pi(\mathbf{s}) \prod_{i} P_i\bigl(s_i \to s_i' \mid u_i(\mathbf{s})\bigr)
  \neq
  \pi(\mathbf{s}') \prod_{i} P_i\bigl(s_i' \to s_i \mid u_i(\mathbf{s}')\bigr) ,
\label{eq:fail}
\end{equation}
and the detailed balance condition~\eqref{eq:db} fails. To preserve detailed balance in parallel Markov chain Monte Carlo, synchronous updates must be restricted to sets of spins that do not interact directly (e.g., checkerboard updates on a bipartite lattice~\cite{heermann1990parallelization}), or be implemented as a proper block update that samples from the exact joint conditional distribution of the updated spins given the remaining spins~\cite{robert1999monte}.

Another practical drawback of naive synchronous all-spin updates is that they can induce oscillatory dynamics. Here, oscillation refers to the Markov chain repeatedly cycling between a small set of configurations instead of mixing over the state space. For example, updating all spins simultaneously based on the previous configuration can cause the system to alternate between two complementary patterns on successive steps. This leads to period-2 dynamics in which the distribution over configurations oscillates between successive steps instead of settling smoothly toward equilibrium, even though a stationary distribution exists. Such oscillations can be mitigated by avoiding fully synchronous all-spin updates, for example by using checkerboard updates on a bipartite lattice~\cite{heermann1990parallelization}.

\subsection{Precision of Coupling Coefficients}
Many combinatorial optimization problems, when formulated as Ising models, give rise to coefficients that are arbitrary integers or real numbers representing costs or weights~\cite{lucas2014ising}. 

\begin{figure}[h]
  \centering
  \vspace{-0.5em}
  \includegraphics[width=\columnwidth]{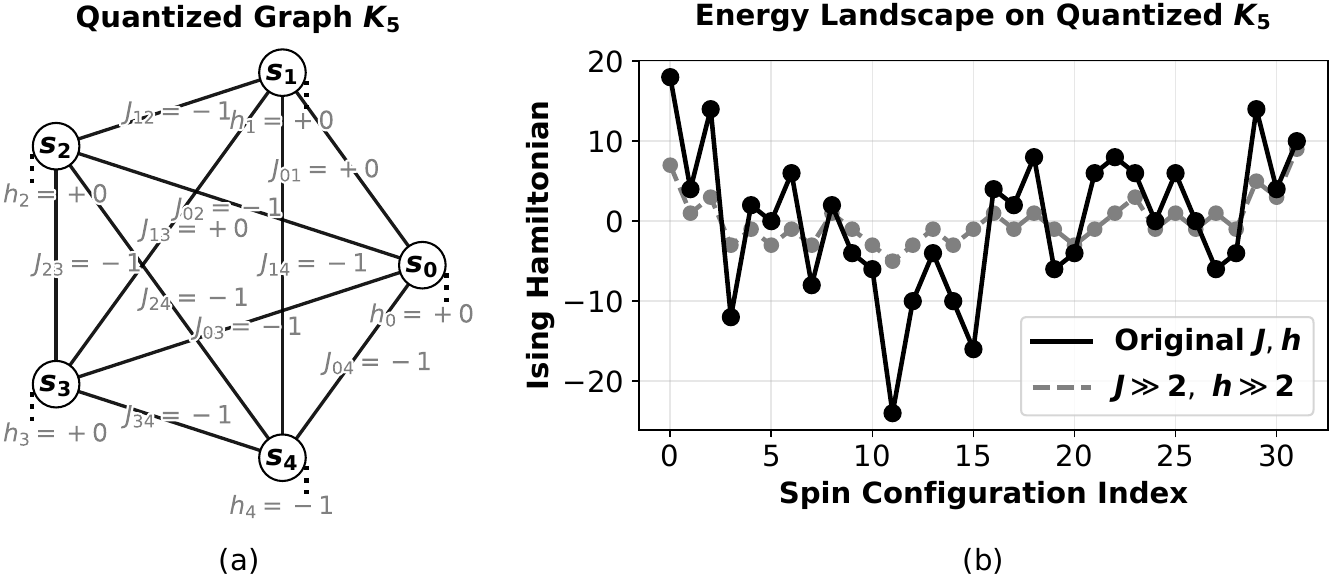}
  \vspace{-2em}
  \caption{(a) The graph \(K_5\) in Figure~\ref{fig:k5} is quantized by a 2-bit arithmetic right shift of the couplings \(J_{ij}\) and external fields \(h_i\). (b) Energy landscape before (solid) and after (dashed) the 2-bit arithmetic right shift.}
  \label{fig:k5_shifted}
  \vspace{-0.5em}
\end{figure}

On hardware that supports only 1-bit couplings, representing these weights often requires either coarse quantization or the introduction of ancilla spins~\cite{9294135}. Coarse quantization distorts the energy landscape (see Figure~\ref{fig:k5_shifted}) and can change the ground state, so the Ising machine no longer solves the original mathematical problem exactly~\cite{9730601, yarkoni2022quantum}. Ancilla-based encodings and related constructions inflate the problem size, increase the effective physical graph connectivity, and waste hardware resources, thereby directly hurting scalability and time-to-solution~\cite{PRXQuantum.2.040322}. Thus, to encode such combinatorial optimization problems exactly onto the hardware topology, Ising machines need to support scalable precision in their coupling coefficients.

However, implementing high-bit-width coupling coefficients is challenging, especially in Ising machines that leverage analog circuitry, such as quantum annealing hardware~\cite{boothby2020next} and analog CMOS Ising machines~\cite{9162869}. For example, in a D-Wave quantum processing unit, the local fields and couplers are realized by analog control circuits driven by digital-to-analog converters, so the programmed Hamiltonian is perturbed by various noise sources such as low-frequency noise, calibration drift, and digital-to-analog discretization errors. These effects appear as random shifts in the effective biases and couplings and can even induce spurious couplings between spins that are not directly connected in the hardware graph. Because these perturbations are non-negligible relative to the programmable range, differences between nominal coupling values below this noise scale cannot be reliably resolved, so simply increasing the bit-width of logical coefficients does not translate into proportionally higher physical precision~\cite{yarkoni2022quantum}.

%% file: sections/snowball.tex
\section{Snowball}\label{sec:snowball}
Based on the above design considerations, this work presents Snowball, a digital, scalable, all-to-all coupled Ising machine with dual-mode Markov chain Monte Carlo spin selection and asynchronous spin updates for fast combinatorial optimization. The following subsections first describe the high-level algorithm and then the corresponding hardware architecture.

\subsection{Algorithm}
Conceptually, each Monte Carlo step can be decomposed into a spin-selection phase and a spin-update phase. In the spin-selection phase, one or more spins are chosen, and their local fields $u_i$ are computed (see the definition below Equation~\eqref{eq:glauber}). In principle, single-spin, block, or all-spin selection schemes are possible, and the selection can be deterministic or random. In the spin-update phase, one or more spins are then updated based on the computed local fields.

The hardware implementation of the parallel Markov chain Monte Carlo spin-selection scheme can consume more instantaneous dynamic power than that of the sequential spin-selection scheme, because multiple candidate flips are evaluated in parallel at each step. However, as discussed in Section~\ref{subsec:topo}, in an all-to-all topology, each spin interacts with all other spins, so selecting a single flip based on a global view of all candidate flip probabilities can reduce the number of Monte Carlo steps required to reach a good solution, improving time-to-solution. Thus, Snowball integrates dual-mode Markov chain Monte Carlo spin selection, allowing the spin-selection mode to be tailored to each combinatorial optimization instance and to balance time-to-solution against energy consumption.

\subsubsection{Sequential Markov Chain Monte Carlo (Random-Scan)}
In Snowball's sequential mode, the spin-selection phase instantiates a random-scan single-spin Markov chain Monte Carlo kernel with asynchronous single-spin updates. At each iteration, a spin index $i \in \{1,\dots,N\}$ is selected uniformly at random, its local field $u_i$ is computed, and a flip $s_i \to -s_i$ is accepted with probability $P_{\mathrm{flip}}(s_i \to -s_i \mid \mathbf{s})$ given by Equation~\eqref{eq:glauber}. Let $\mathbf{s}^{(i\to -i)}$ denote the configuration obtained from $\mathbf{s}$ by flipping spin $i$, and let $\pi_T(\mathbf{s}) \propto \exp(-H(\mathbf{s})/T)$ denote the Gibbs distribution at temperature $T>0$. The resulting transition kernel $P_{\mathrm{seq}}$ is
\begin{equation}
\label{eq:pseq}
\resizebox{\columnwidth}{!}{$
P_{\mathrm{seq}}(\mathbf{s} \to \mathbf{s}')
=
\begin{cases}
\frac{1}{N}\, P_{\mathrm{flip}}(s_i \to -s_i \mid \mathbf{s}), 
  & \mathbf{s}' = \mathbf{s}^{(i\to -i)}, \\[0.3em]
1 - \sum_{i=1}^N \frac{1}{N}\, P_{\mathrm{flip}}(s_i \to -s_i \mid \mathbf{s}), 
  & \mathbf{s}' = \mathbf{s}, \\[0.3em]
0, 
  & \text{otherwise}.
\end{cases}
$}
\end{equation}
For a neighboring pair $\mathbf{s}$ and $\mathbf{s}' = \mathbf{s}^{(i\to -i)}$, write $\Delta E_i(\mathbf{s}) \equiv H(\mathbf{s}^{(i\to -i)}) - H(\mathbf{s})$. By definition of the Gibbs distribution and Equation~\eqref{eq:glauber},
\begin{equation}
\begin{aligned}
\pi_T(\mathbf{s})
&\propto \exp\bigl(-H(\mathbf{s})/T\bigr), \\
\pi_T(\mathbf{s}^{(i\to -i)})
&\propto \exp\bigl(-(H(\mathbf{s})+\Delta E_i(\mathbf{s}))/T\bigr), \\
P_{\mathrm{flip}}(s_i \to -s_i \mid \mathbf{s})
&= \frac{1}{1+\exp(\Delta E_i(\mathbf{s})/T)}, \\
P_{\mathrm{flip}}(s_i \to -s_i \mid \mathbf{s}^{(i\to -i)})
&= \frac{1}{1+\exp(-\Delta E_i(\mathbf{s})/T)}.
\end{aligned}
\end{equation}
A straightforward calculation then shows
\begin{equation}
\resizebox{\columnwidth}{!}{$
\pi_T(\mathbf{s})\, P_{\mathrm{flip}}(s_i \to -s_i \mid \mathbf{s})
=
\pi_T(\mathbf{s}^{(i\to -i)})\, P_{\mathrm{flip}}(s_i \to -s_i \mid \mathbf{s}^{(i\to -i)})
$.}
\end{equation}
Multiplying both sides by $1/N$ yields
\begin{equation}
\resizebox{\columnwidth}{!}{$
\pi_T(\mathbf{s})\, P_{\mathrm{seq}}(\mathbf{s} \to \mathbf{s}^{(i\to -i)})
=
\pi_T(\mathbf{s}^{(i\to -i)})\, P_{\mathrm{seq}}(\mathbf{s}^{(i\to -i)} \to \mathbf{s})
$,}
\end{equation}
so the sequential kernel $P_{\mathrm{seq}}$ satisfies detailed balance with respect to the Gibbs distribution $\pi_T$. Summing over all $\mathbf{s}$ shows that $\pi_T$ is a stationary distribution of the Markov chain.

Moreover, for any finite $T>0$, one has $0 < P_{\mathrm{flip}}(s_i \to -s_i \mid \mathbf{s}) < 1$ for all spins $i$ and configurations $\mathbf{s}$. Thus, from any configuration $\mathbf{s}$, there is a positive probability of flipping any single spin in a finite number of steps, which implies irreducibility on the finite state space. In addition, at each step there is a nonzero probability of remaining in the same configuration (the case where the proposed flip is rejected), so the chain is aperiodic. On a finite state space, irreducibility and aperiodicity together imply that the Markov chain is ergodic and admits a unique stationary distribution, and the detailed balance condition guarantees that this stationary distribution is exactly the Gibbs distribution $\pi_T$. Consequently, Snowball's sequential Markov chain Monte Carlo spin-selection mode with asynchronous single-spin updates converges to the target Gibbs distribution at fixed temperature, and time averages of observables converge to their expectations under this distribution. In particular, the presence of self-loops (rejected flips) prevents the period-2 oscillations characteristic of naive fully synchronous update schemes.

\subsubsection{Parallel Markov Chain Monte Carlo (Roulette-Wheel)}
To accelerate time-to-solution, parallelism can be leveraged. However, as shown in Equation~\eqref{eq:fail}, naive synchronous multi-spin updates in a single Markov chain can violate detailed balance with respect to the Gibbs distribution. Additionally, as discussed in Section~\ref{subsec:conv}, naive all-spin updates can induce oscillatory behavior due to period-2 dynamics, in which the distribution over configurations alternates between successive steps instead of settling smoothly toward equilibrium. 

\begin{figure}[h]
  \centering
  \vspace{-1.5em}
  \includegraphics[width=\columnwidth]{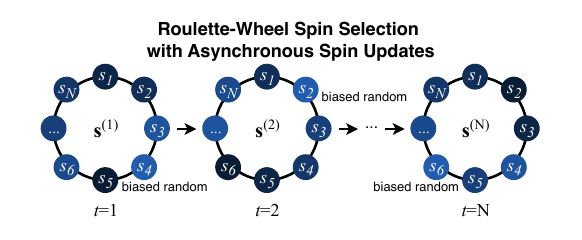}
  \vspace{-3em}
  \caption{Roulette-wheel spin selection with asynchronous single-spin updates. The highlighted spin is selected by a biased random rule, where spin $i$ is sampled with probability proportional to its candidate flip probability $P_{\mathrm{flip}}(s_i \to -s_i \mid \mathbf{s})$.}
  \label{fig:bmc}
  \vspace{-0.5em}
\end{figure}

In Snowball's parallel mode, the spin-selection phase instead instantiates a biased random single-spin Markov chain Monte Carlo kernel with asynchronous single-spin updates: candidate flips for all spins are evaluated in parallel, but at most a single spin is selected based on its weight given by the normalized flip probability (roulette-wheel spin selection), and then updated asynchronously per step. This is illustrated in Figure~\ref{fig:bmc}. 

For the current configuration $\mathbf{s}$ and temperature $T$, the local fields $u_i$ are computed for all $i$, and the corresponding single-spin flip probabilities $P_{\mathrm{flip}}(s_i \to -s_i \mid \mathbf{s})$ are obtained from Equation~\eqref{eq:glauber}. These probabilities are then interpreted as weights in a categorical distribution over spin indices as follows:
\begin{equation}
\Pr(I = i \mid \mathbf{s}) =
\frac{P_{\mathrm{flip}}(s_i \to -s_i \mid \mathbf{s})}
     {\sum_{j=1}^N P_{\mathrm{flip}}(s_j \to -s_j \mid \mathbf{s})},
\quad i = 1,\dots,N,
\label{eq:weight}
\end{equation} where $I$ denotes the random index of the spin selected for update and $0 < \sum_{j=1}^N P_{\mathrm{flip}}(s_j \to -s_j \mid \mathbf{s}) < \infty$. Equivalently, a single spin is sampled with probability proportional to its candidate flip probability and is then deterministically flipped, while all other spins remain unchanged, hence the term roulette-wheel spin selection. If the aggregate flip probability becomes numerically degenerate (e.g., vanishing or non-finite total weight), Snowball falls back to a conventional one-site update with random-scan single-spin selection and asynchronous single-spin updates. 

While the spin selection is stochastic, the spin update itself is deterministic once the index is chosen, so the kernel is rejection-free at the level of single-spin updates. Since exactly one spin is flipped at every step and there are no self-loops in this kernel, the resulting Markov chain has period-2: each configuration can only be revisited after an even number of steps, and the distributions at even and odd time steps can oscillate.

Nevertheless, for an irreducible Markov chain on a finite state space, such periodicity does not preclude the existence of a unique stationary distribution, and the ergodic theorem guarantees that time averages of observables along the chain converge to their expectations under this stationary distribution. Intuitively, periodicity does not prevent the long-run empirical averages from settling to the stationary expectation, because every state is visited with the correct asymptotic frequency in the irreducible finite-state setting. Moreover, this deterministic spin flip based on Equation~\eqref{eq:weight} reduces spin-update rejection, accelerates the discovery of low-energy configurations, and improves time-to-solution on combinatorial optimization instances~\cite{nambu2022rejection}. 

Algorithm~\ref{alg:dms} presents pseudocode for Snowball’s dual-mode spin selection and asynchronous spin updates. Note that simulated annealing is applied only for $T > 0$. Simulated annealing modifies the Markov chain Monte Carlo dynamics by lowering the effective temperature during the run. At high temperatures, the chain explores the configuration space broadly and can escape local minima, while at low temperatures it concentrates around low-energy states. Although this time-inhomogeneous process does not converge to a fixed Gibbs distribution, an appropriately chosen cooling schedule can improve convergence to low-energy (near-optimal) configurations and thus reduce the time-to-solution for optimization tasks.

Another common annealing mechanism is parallel tempering (replica-exchange Markov chain Monte Carlo)~\cite{geyer1991markov}, in which multiple replicas at different temperatures are simulated in parallel and occasional exchanges of spin configurations between neighboring temperatures, accepted with a suitable swap-acceptance probability, accelerate barrier crossing in rough landscapes. However, as system size or heat capacity increases, maintaining adequate swap acceptance typically requires many closely spaced temperature replicas; otherwise, swap acceptance declines and mixing degrades~\cite{kone2005selection, nadler2007dynamics}. For these reasons, this work focuses on simulated annealing.

After a spin flip, Snowball uses incremental updates of the local fields rather than recomputing them from scratch. Recall that the local field at spin $i$ is
\begin{equation}
u_i = h_i + \sum_{j \neq i} J_{ij} s_j.
\end{equation}
If a single spin $k$ is flipped from $s_k^{\text{old}}$ to $s_k^{\text{new}} (= -s_k^{\text{old}})$, then the contribution of spin $k$ to the local field of any other spin $i \neq k$ changes from $J_{ik} s_k^{\text{old}}$ to $J_{ik} s_k^{\text{new}} (= -J_{ik} s_k^{\text{old}})$. Thus, the updated local field is
\begin{equation}
u_i'
= u_i - J_{ik} s_k^{\text{old}} + J_{ik} s_k^{\text{new}}
= u_i - 2 J_{ik} s_k^{\text{old}}.
\end{equation}
This relation allows each affected local field to be updated by a simple constant-time operation rather than recomputing the full sum over all neighbors. For an $N$-spin all-to-all topology, recomputing all local fields from scratch after each accepted spin flip requires $\Theta(N)$ work for each of the $N$ local fields, for a total cost of $\Theta(N^2)$ per accepted flip. With incremental updates, a flip of a single spin affects only the $N-1$ other local fields, so exactly those fields are updated, incurring a computational cost of $\Theta(N)$ per accepted flip.

\begin{algorithm}[h]
  \caption{\small Dual-Mode Selection and Asynchronous Updates}
  \label{alg:dms}
  \begin{algorithmic}[1]\small
    \Require Graph $G$ (couplings $J_{ij}$, fields $h_i$), number of spins $N$, number of iterations $K$,
            initial configuration $\mathbf{s}^{(1)}$, initial and final temperatures $T_0, T_1$
    \State $\mathbf{s} \gets \mathbf{s}^{(1)}$
    
    \For{$i = 1,\dots,N$} 
      \State $u_i \gets h_i + \sum_{j \ne i} J_{ij} s_j$ \Comment{Local-field initialization}
    \EndFor
    
    \For{$t = 1,\dots,K$}
      \State $T \gets \textsc{Cooling}(T_0, T_1, t, K)$ \Comment{Annealing if $T>0$}
    
      \For{$i = 1,\dots,N$ \textbf{in parallel}}
        \State $\Delta E_i \gets 2 s_i u_i$
        \State $p_i \gets P_{\mathrm{flip}}(\Delta E_i)$
      \EndFor
      
      \If{$\sum_{i=1}^N p_i \le 0$ \textbf{or} $\sum_{i=1}^N p_i$ is not finite}
        \State Draw $i \sim \text{Uniform}(\{1,\dots,N\})$ \Comment{Random-scan}
        \State $\Delta E_i \gets 2 s_i u_i$
        \State $p_i \gets P_{\mathrm{flip}}(\Delta E_i)$
        \State Draw $r \sim \text{Uniform}(0,1)$
        \If{$r < p_i$}
          \State $s^{\text{old}} \gets s_i$ \Comment{Pre-flip spin cache}
          \State $s_i \gets -s_i$
          \For{$j = 1,\dots,N$} 
            \If{$j \ne i$}
              \State $u_j \gets u_j - 2 J_{j i}\,s^{\text{old}}$ \Comment{Async. incr. update}
            \EndIf
          \EndFor
        \EndIf
        
      \Else \Comment{Roulette-wheel}
        \State Draw $I \sim \text{Categorical}(\{\, p_i / \sum_{j=1}^N p_j \,\}_{i=1}^N)$
        \State $s^{\text{old}} \gets s_I$ \Comment{Pre-flip spin cache}
        \State $s_I \gets -s_I$
        \For{$j = 1,\dots,N$} 
          \If{$j \ne I$}
            \State $u_j \gets u_j - 2 J_{j I}\,s^{\text{old}}$ \Comment{Async. incr. update}
          \EndIf
        \EndFor
      \EndIf
    \EndFor

    \State \Return final configuration $\mathbf{s}$
  \end{algorithmic}
\end{algorithm}

\subsection{Hardware Architecture}\label{subsec:arch}
\begin{figure*}[t]
  \centering
  \includegraphics[width=\textwidth]{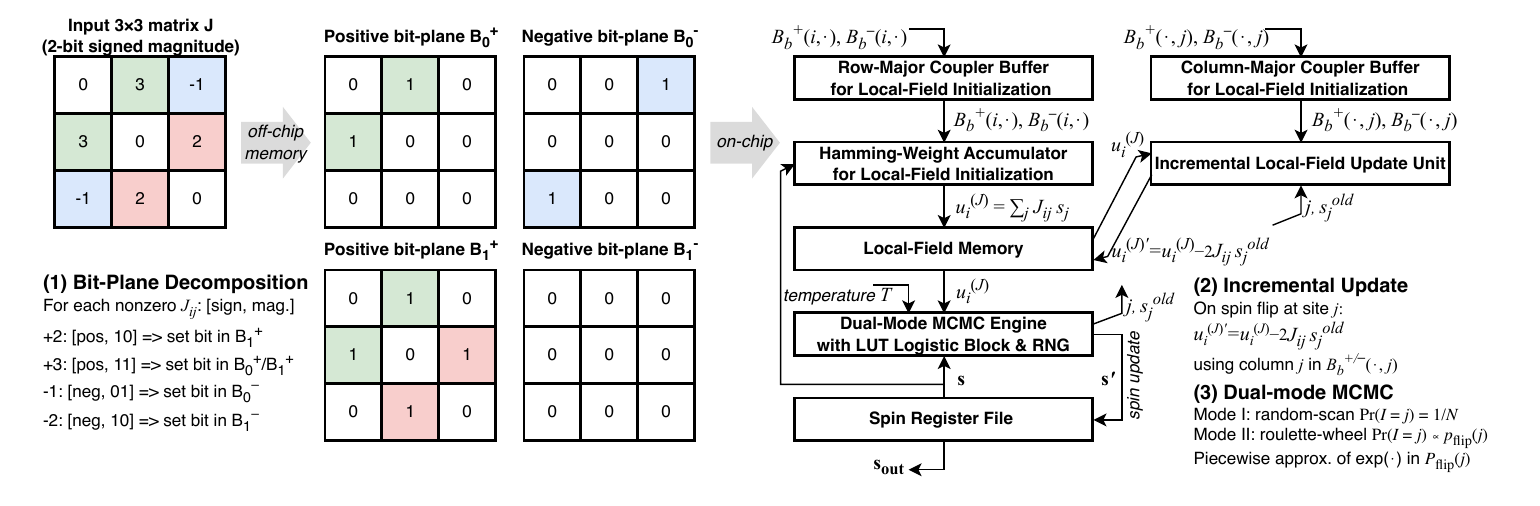}
  \vspace{-2.5em}
  \caption{Hardware architecture of Snowball. The input matrix \(J\) is stored off-chip in positive/negative bit-planes \(B_b^{\pm}\), where a 2-bit signed-magnitude coupling corresponds to \(B = 2\) magnitude bit-planes. In real implementation, coupler bits are packed into 64-bit words and streamed into row-major on-chip tile buffers (BRAM) to initialize the coupler-induced local fields \(u_i^{\scriptscriptstyle(J)} = \sum_j J_{ij} s_j\) via a Hamming-weight accumulator. The local fields \(u_i^{\scriptscriptstyle(J)}\) and external biases \(h_i\) are stored in small on-chip memories, and the full local field used by the MCMC engine is \(u_i = u_i^{\scriptscriptstyle(J)} + h_i\). During sampling, the dual-mode MCMC engine (random-scan and roulette-wheel) uses a LUT-based piecewise approximation of the exponential to evaluate flip probabilities at temperature \(T\), driven by an on-chip stateless RNG. After each accepted spin flip \(s_j\), the local fields are updated incrementally by streaming column \(j\) of the column-major bit-planes and applying a read-modify-write update \(u_i^{\scriptscriptstyle(J)} \leftarrow u_i^{\scriptscriptstyle(J)} - 2 J_{ij} s_j^{\text{old}}\) to the on-chip local-field memory.}
  \vspace{-1.0em}
  \label{fig:arch}
\end{figure*}

Snowball implements an Ising model over \(N\) binary spins \(s_i \in \{-1,+1\}\) for \(i = 0,\dots,N-1\), with Hamiltonian \(H(\mathbf{s}) = -\frac{1}{2} \sum_{i,j=0}^{N-1} J_{ij} s_i s_j - \sum_{i=0}^{N-1} h_i s_i\). On hardware, spins are encoded as bits \(x_i = (s_i + 1)\big/2 \in \{0,1\}\), with \(s_i = 2 x_i - 1\), and packed into \(W = N/64\) 64-bit words. The architecture is built around three key ideas:
\begin{enumerate}[leftmargin=*,topsep=0pt,itemsep=0pt]
\item A multi-bit bit-plane representation of the dense coupling matrix \(J\) in both row-major and column-major form.
\item On-chip storage of the coupler-induced local fields \(u_i^{(J)} = \sum_j J_{ij} s_j\), updated incrementally after each spin flip.
\item A shared stochastic update engine that supports two Markov chain Monte Carlo (MCMC) modes (random-scan selection and roulette-wheel selection) and implements a programmable simulated annealing schedule over the temperature \(T\).
\end{enumerate}
These design choices are motivated by hardware constraints (off-chip bandwidth, BRAM capacity, logic utilization) and are precisely what make scalable all-to-all connectivity, dual-mode MCMC with simulated annealing, and high-precision coupling coefficients feasible within an FPGA implementation. Figure~\ref{fig:arch} illustrates the overall hardware architecture of Snowball.

\subsubsection{Multi-Bit Bit-Plane Representation of Couplings}\label{subsubsec:bitp}
To support high-precision couplings while keeping memory and datapath widths manageable, the coupler matrix \(J\) is represented using signed bit-planes. For each pair \((i,j)\),
\begin{equation}
    J_{ij}
    = \sum_{b=0}^{B-1} 2^b \bigl( B_b^+(i,j) - B_b^-(i,j) \bigr),
    \label{eq:J-bitplanes-arch}
\end{equation}
where \(B_b^+(i,j), B_b^-(i,j) \in \{0,1\}\) indicate positive and negative contributions on bit-plane \(b\), respectively. The number of active planes \(B\) directly controls the available dynamic range and resolution of \(J_{ij}\).

For each bit-plane $b$, the design maintains $B_b^+, B_b^- \in \{0,1\}^{N \times N}$ in row-major layout and $B_b^{+,\top}, B_b^{-,\top}$ in column-major layout. This separation of sign and magnitude, together with the dual layouts, is fundamental to the scalability of the architecture for two reasons: (a) bit-planes allow storing and processing couplings using 1-bit values per plane, so increasing precision only grows memory linearly in $B$, without widening the internal arithmetic beyond what is needed for the accumulated fields; and (b) row-major and column-major layouts enable streaming-friendly access patterns for dense initialization and efficient incremental updates, respectively.

\subsubsection{Local Field Representation and Incremental Updates}
For sampler updates, the relevant quantity is the local field at each site, $u_i^{(J)} = \sum_{j=0}^{N-1} J_{ij} s_j$ and $u_i = u_i^{(J)} + h_i$. The architecture stores \(u_i^{(J)}\) and \(h_i\) in on-chip BRAM. The full vector \(\mathbf{u}^{(J)}\) is first initialized from scratch using the row-major bit-planes and then updated incrementally after each single-spin update using the column-major bit-planes.

\paragraph{Initialization with row-major bit-planes}
Let \(W = N/64\) be the number of 64-spin words. For a given bit-plane \(b\) with the weight \(w_b = 2^b\), row \(i\), and word index \(w\), define
\begin{align}
    m_P &= \sum_{j} B_b^+(i,j), &
    o_P &= \sum_{j} B_b^+(i,j)\, x_j, \\
    m_N &= \sum_{j} B_b^-(i,j), &
    o_N &= \sum_{j} B_b^-(i,j)\, x_j,
\end{align}
where \(x_j = (s_j + 1)/2 \in \{0,1\}\) encodes the spins, and the sums are taken over the 64 column indices whose couplers are packed into the \(w\)-th 64-bit word. Here, \(m_P\) and \(m_N\) are the Hamming weights (number of 1-bits) of the corresponding 64-bit positive and negative coupler words, while \(o_P\) and \(o_N\) are the Hamming weights of the bitwise AND between these coupler words and the 64-bit spin words.

Among the neighbors with $B_b^+(i,j)=1$ in this word, there are $o_P$ spins with $s_j = +1$ and $m_P - o_P$ with $s_j = -1$, so $\sum_{j:B_b^+(i,j)=1} s_j = 2 o_P - m_P$. The contribution of these positive couplers in bit-plane $b$ and word $w$ to $u_i^{(J)}$ is therefore $\Delta u_i^{(J,+)}(b,w) = w_b (2 o_P - m_P)$. Similarly, for the negative couplers, $\Delta u_i^{(J,-)}(b,w) = - w_b (2 o_N - m_N)$. Summing over all bit-planes and words yields
\begin{equation}
u_i^{(J)} = \sum_{b=0}^{B-1} \sum_{w=0}^{W-1} \bigl( \Delta u_i^{(J,+)}(b,w) + \Delta u_i^{(J,-)}(b,w) \bigr) = \sum_{j=0}^{N-1} J_{ij} s_j .
\label{eq:local-field-bits}
\end{equation} This Hamming-weight-based accumulation is highly parallelizable and uses only bitwise operations and integer adds, which are efficient on FPGA fabric. It supports dense all-to-all connectivity without requiring explicit \(N^2\) multipliers.

\paragraph{Incremental updates with column-major bit-planes}
When a single spin \(s_j\) flips, the coupler-induced fields change as
\begin{equation}
\resizebox{\columnwidth}{!}{$
    u_i^{(J)\,\prime}
    = \sum_k J_{ik} s_k'
    = u_i^{(J)} + J_{ij}\bigl(s_j^{\text{new}} - s_j^{\text{old}}\bigr)
    = u_i^{(J)} - 2 J_{ij} s_j^{\text{old}}
$.}
\label{eq:u-increment-basic}
\end{equation}
Substituting Equation~\eqref{eq:J-bitplanes-arch} gives
\begin{equation}
\resizebox{\columnwidth}{!}{$
    \Delta u_i^{(J)}
    = u_i^{(J)\,\prime} - u_i^{(J)}
    = -2 s_j^{\text{old}} \sum_{b=0}^{B-1} 2^b
      \bigl( B_b^+(i,j) - B_b^-(i,j) \bigr)
$.}
\end{equation}
In hardware, the column-major bit-planes \(B_b^{+,\top}\) and \(B_b^{-,\top}\) are scanned for column \(j\); for each row index \(i\) where a bit is set:
\begin{align}
    \text{if } B_b^{+,\top}(j,i) = 1 &: 
    \quad u_i^{(J)} \leftarrow u_i^{(J)} - 2 \cdot 2^b \, s_j^{\text{old}}, \\
    \text{if } B_b^{-,\top}(j,i) = 1 &: 
    \quad u_i^{(J)} \leftarrow u_i^{(J)} + 2 \cdot 2^b \, s_j^{\text{old}}.
\end{align}
This realizes exact incremental updates for a dense all-to-all topology using only integer additions and bit scanning, making it practical to maintain fully connected couplings at scale. This incremental maintenance of \(\{u_i^{(J)}\}\) avoids recomputing \(\sum_{j} J_{ij} s_j\) from scratch at every step and instead updates all local fields with a single pass over the column \(J_{\cdot j}\), reducing the per-update cost from a dense matrix-vector product to a sequence of integer additions that maps efficiently to the FPGA fabric even under all-to-all connectivity.

\subsubsection{Dual-Mode MCMC Engine with Simulated Annealing}\label{subsubsec:hw}
The kernel supports two MCMC schemes over the same datapath: a random-scan selection mode and a roulette-wheel selection mode. Both can be driven by a programmable simulated annealing schedule \(\{T_k\}_{k=0}^{K-1}\) over the temperature \(T\).

\paragraph{Local flip probability and piecewise approximation}
Given the current configuration \(\mathbf{s}\), the local field at site \(i\) is \(u_i(\mathbf{s}) \equiv h_i + \sum_{j \neq i} J_{ij} s_j\), consistent with Equation~\eqref{eq:glauber}. In hardware this is decomposed as $u_i^{(J)} \equiv \sum_{j\neq i} J_{ij} s_j$ and $u_i(\mathbf{s}) = u_i^{(J)} + h_i,$ where \(u_i^{(J)}\) is stored on chip and maintained incrementally after each spin flip as described in the previous section. For a proposed flip \(s_i \to -s_i\), the corresponding energy change is \(\Delta E_i \equiv H(\mathbf{s}^{(i\to -i)}) - H(\mathbf{s}) = 2 s_i u_i(\mathbf{s})\), and the target Glauber flip probability is \(P_{\mathrm{flip}}(s_i \to -s_i \mid \mathbf{s}) = (1 + \exp(\Delta E_i / T))^{-1}\), as in Equation~\eqref{eq:glauber}.

On hardware, the temperature \(T\) is represented in fixed-point format, and the quantity \(z_i = \Delta E_i / T = 2 s_i u_i(\mathbf{s}) / T\) is mapped through a piecewise-linear lookup table, yielding an approximation
\begin{equation}
p_{\mathrm{flip}}(i) \approx P_{\mathrm{flip}}(s_i \to -s_i \mid \mathbf{s}),
\end{equation}
which is shared by both update modes. This piecewise-linear approximation replaces the computationally expensive exponential in \(\exp(\Delta E_i/T)\) with table lookups and fixed-point arithmetic, substantially reducing latency and resource usage while still providing sufficient numerical accuracy for the MCMC dynamics.

\paragraph{Mode~I: random-scan selection}
In the random-scan selection mode, each iteration performs a Glauber update on a randomly chosen spin:

\noindent\textbf{Site selection.}
A stateless pseudo-random generator produces a uniform integer
\(u \in \{0,\dots,2^{32}-1\}\). The site index
\begin{equation}
    j = \left\lfloor 
        \frac{u \, N}{2^{32}}
    \right\rfloor
\end{equation}
is thus uniformly distributed over \(\{0,\dots,N-1\}\).

\noindent\textbf{Local field and energy change.}
The coupler field \(u_j^{(J)}\) and bias \(h_j\) are read from BRAM, and
\begin{equation}
    u_j(\mathbf{s}) = u_j^{(J)} + h_j
\end{equation}
is formed. The flip energy change is then
\begin{equation}
    \Delta E_j = 2 s_j u_j(\mathbf{s}).
\end{equation}
This \(\Delta E_j\) is passed to the logistic block to obtain
\begin{equation}
    p_{\mathrm{flip},j} \approx
    \frac{1}{1 + \exp(\Delta E_j / T)}.
\end{equation}

\noindent\textbf{Random-scan flip decision.}
A second uniform random integer \(v \in \{0,\dots,2^{32}-1\}\) is drawn and rescaled to a uniform real in \([0,1)\). The proposed flip \(s_j \to -s_j\) is accepted with probability \(p_{\mathrm{flip},j}\), i.e.,
\begin{equation}
    s_j^{\text{new}} =
    \begin{cases}
        -s_j^{\text{old}}, & \text{if } \mathrm{Unif}[0,1) < p_{\mathrm{flip},j},\\[2pt]
        s_j^{\text{old}},  & \text{otherwise}.
    \end{cases}
\end{equation}

\noindent\textbf{Asynchronous field update.}
If a flip is accepted (\(s_j^{\text{new}} \neq s_j^{\text{old}}\)), the incremental rule
\begin{equation}
    u_i^{(J)} \leftarrow u_i^{(J)} - 2 J_{ij} s_j^{\text{old}},
    \quad \forall i
\end{equation}
is applied using the column-major bit-planes, as in Equation~\eqref{eq:u-increment-basic}.

Exactly one spin is updated per iteration, and its influence is immediately reflected in all local fields, realizing an asynchronous random-scan Glauber dynamics. Simulated annealing is implemented by varying \(T\) across iterations according to the preloaded schedule \(\{T_k\}\).

\paragraph{Mode~II: roulette-wheel selection}
In the roulette-wheel selection mode, each iteration evaluates flip probabilities for all spins and then selects at most one spin to flip using a roulette-wheel rule.

\noindent\textbf{Per-site flip probability.}
For each site \(i\), the local field \(u_i(\mathbf{s})\) and energy change \(\Delta E_i = 2 s_i u_i(\mathbf{s})\) are computed as above, and the Glauber flip probability \(p_{\mathrm{flip}}(i) \equiv P_{\mathrm{flip}}(s_i \to -s_i \mid \mathbf{s}) \approx (1 + \exp(\Delta E_i / T))^{-1}\) is obtained from the lookup table.

\noindent\textbf{Global weight and roulette-wheel selection.}
The architecture accumulates
\begin{equation}
    W = \sum_{i=0}^{N-1} p_{\mathrm{flip}}(i).
\end{equation}
If \(W = 0\), no site has a nonzero flip probability. In this case, the kernel
falls back to a sequential mode with random-scan spin selection and asynchronous
spin updates. Otherwise, a random number \(r \in [0,W)\) is drawn, and the
unique site \(j\) satisfying $\sum_{i=0}^{j-1} p_{\mathrm{flip}}(i) \le r < \sum_{i=0}^{j} p_{\mathrm{flip}}(i)$ is selected. Since \(p_{\mathrm{flip}}(i)\) is the hardware approximation of
\(P_{\mathrm{flip}}(s_i \to -s_i \mid \mathbf{s})\), the roulette-wheel
selection implements
\begin{align}
\Pr(I = j \mid \mathbf{s})
&= \frac{p_{\mathrm{flip}}(j)}{\sum_{k=1}^N p_{\mathrm{flip}}(k)} \\[4pt]
&\approx
\frac{P_{\mathrm{flip}}(s_j \to -s_j \mid \mathbf{s})}
     {\sum_{k=1}^N P_{\mathrm{flip}}(s_k \to -s_k \mid \mathbf{s})}.
\label{eq:weight-hw}
\end{align}
This is the discrete selection rule in Equation~\eqref{eq:weight}. In an optional uniformized variant, \(W\) is compared against a fixed maximum
rate \(W^{*} = N\), and with probability \(1 - W / W^{*}\) no flip is performed,
yielding a uniformized continuous-time Markov chain~\cite{stewart2021introduction}. With uniformization
enabled, when \(W = 0\) the iteration is always a null transition (no spin is
flipped).

\noindent\textbf{Deterministic flip and asynchronous update.}
Once site \(j\) is selected, the spin is flipped deterministically, $s_j^{\text{new}} = - s_j^{\text{old}},$ and the coupler fields are updated via
\begin{equation}
    u_i^{(J)} \leftarrow u_i^{(J)} - 2 J_{ij} s_j^{\text{old}},
    \quad \forall i.
\end{equation}
As in random-scan selection mode, only one spin is updated per iteration, and its effect on all local fields is propagated immediately, preserving the asynchronous single-spin update semantics.

\paragraph{Stateless RNG}
Snowball employs a stateless pseudorandom number generator for both random-scan and roulette-wheel selection. Each variate is computed as a pure function of a global 64-bit seed supplied by the host and a small set of indices (e.g., annealing stage~$k$, iteration~$t$, and a purpose-specific salt~$r$), rather than by updating a global RNG state. This stateless design offers two advantages on FPGAs: (i) independent random numbers can be generated in parallel by varying~$r$, without contention on shared RNG state; and (ii) the mixing function maps efficiently to FPGA LUTs and DSPs, eliminating the need for large state machines or complex RNG cores.

%% file: sections/evaluation.tex
\section{Evaluation}\label{sec:evaluation}
The goal of this evaluation is twofold: first, to assess the solution quality of Snowball’s algorithm on the standard combinatorial optimization benchmark suite Gset~\cite{gset}; and second, to demonstrate its time-to-solution on the K2000 Max-Cut instance on real hardware, in comparison with prior works, including Neal~\cite{dwave_neal}, CIM~\cite{doi:10.1126/science.aah4243}, SB~\cite{goto2019combinatorial}, STATICA~\cite{9062965}, and ReAIM~\cite{10609617}. Among these, ReAIM~\cite{10609617} represents the current state of the art in Ising machines and is therefore used as the primary evaluation baseline.

\subsection{Experimental Methodology}
\subsubsection{Hardware Platform}
\begin{figure}[h]
  \centering
  \vspace{-2em}
  \includegraphics[width=0.9\columnwidth]{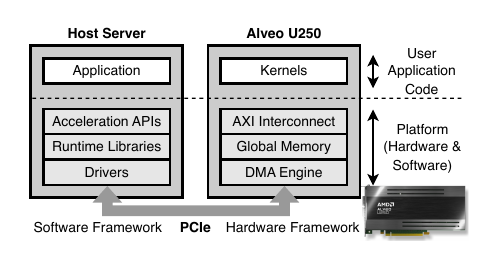}
  \vspace{-2em}
  \caption{Hardware-software stack for an AMD Alveo U250 FPGA accelerator card, illustrating host-side software, PCIe communication, and the on-board accelerator framework.}
  \label{fig:alveo}
  \vspace{-0.5em}
\end{figure}

Figure~\ref{fig:alveo} illustrates the hardware-software stack used to implement the Ising machine on an AMD Alveo U250 FPGA accelerator card. The application executes on the host server and invokes runtime libraries and device drivers to configure the card and manage data transfers. Host-device communication occurs over the PCIe interface to the U250 platform, where an AXI interconnect links on-board global memory, the DMA engine, and the FPGA kernels.

\subsubsection{Benchmarks}
Table~\ref{tab:gset} summarizes the Gset~\cite{gset} benchmark instances used in this experiment. Here, $|V|$ and $|E|$ denote the numbers of vertices and edges; $|E^{+}|$ and $|E^{-}|$ denote the numbers of $+1$ and $-1$ edges; and $\rho = 2|E|/(|V|(|V|-1))$ is the edge density. The values of $\rho$ indicate that Gset~\cite{gset} instances are relatively sparse graphs, which are generally easier to solve. To provide a more challenging benchmark on real hardware, an additional instance $K_{2000}$ is constructed as a complete graph with $|V|=2000$ and couplings $J_{ij} \in \{-1,+1\}$ drawn uniformly at random.
 
\begin{table}[h]
  \centering
  \caption{Summary of Gset~\cite{gset} Benchmark Instances and $K_{2000}$}
  \label{tab:gset}
  \footnotesize
  \setlength{\tabcolsep}{5.3pt}
  \begin{tabularx}{\columnwidth}{@{}l l r r r r r@{}}
    \toprule
    Instance & Topology & $|V|$ & $|E|$ & $|E^{+}|$ & $|E^{-}|$ & $\rho$ \\
    \midrule
    G6        & Erd\H{o}s-R\'enyi & 800     & 19176      & 9665       & 9511       & 6.0\% \\
    G61       & Erd\H{o}s-R\'enyi & 7000 & 17148      & 8755       & 8393       & 0.1\% \\
    G18       & Small-world       & 800     & 4694       & 2379       & 2315       & 1.5\% \\
    G64       & Small-world       & 7000 & 41459      & 20993      & 20466      & 0.2\% \\
    G11       & Torus             & 800     & 1600       & 817           & 783           & 0.5\% \\
    G62       & Torus             & 7000 & 14000      & 6960       & 7040       & 0.1\% \\
    K2000     & Complete          & 2000 & 1999000 & 998314     & 1000686 & 100.0\% \\
    \bottomrule
  \end{tabularx}
\vspace{-2em}
\end{table}

\subsection{Experimental Results}
\subsubsection{Solution Quality on Gset}
For a fair comparison, all algorithms benchmarked in ReAIM~\cite{10609617}, together with Snowball’s RSA (random-scan spin selection with asynchronous spin updates) and RWA (roulette-wheel spin selection with asynchronous spin updates), are reimplemented following the original descriptions and parameter settings. Some parameter values are not specified, so the results for certain instances may differ from those reported in the original paper. 

\begin{table}[h]
\vspace{-1em}
  \centering
  \caption{Comparison of Solution Quality on Gset~\cite{gset} Max-Cut Instances (Metric: Cut Value; Higher Is Better)}
  \label{tab:gsetmcp}
  \footnotesize
  \setlength{\tabcolsep}{0.8pt}
  \begin{tabularx}{\columnwidth}{@{}l r r r r r r r r r r r@{}}
    \toprule
          & SFG  & MFG  & SFA     & MFA    & ASF   & AMF   & ASA & Neal & Tabu & \textbf{RWA} & \textbf{RSA} \\
    \midrule
    G6    & 9816 & 9799 & 9816    & 9857   & 9810  & 9868  & 9850  & 9644  & 9644  & \textbf{11545} & \textbf{11335} \\ 
    G61   & 8784 & 8901 & 8793    & 8806   & 8783  & 8905  & 8864  & 8839  & 8733  & \textbf{12901} & \textbf{10845} \\ 
    G18   & 2442 & 2439 & 2451    & 2460   & 2479  & 2460  & 2469  & 2343  & 2352  & \textbf{3034} & \textbf{2970}  \\ 
    G64   & 21104 & 21035 & 21033 & 21131  & 21144 & 21222 & 21091 & 20947 & 20993 & \textbf{25748} & \textbf{23706} \\ 
    G11   & 864 & 880   & 862     & 868    & 866   & 864   & 872   & 836   & 824   & \textbf{1584} & \textbf{1408}  \\  
    G62   & 7158 & 7240 & 7182    & 7128   & 7240  & 7120  & 7148  & 6998  & 7032  & \textbf{11266} & \textbf{9044}  \\  
    \bottomrule
  \end{tabularx}
\vspace{-0.2em}
\end{table}

\begin{figure}[h]
  \centering
  \vspace{-1.8em}
  \includegraphics[width=\columnwidth]{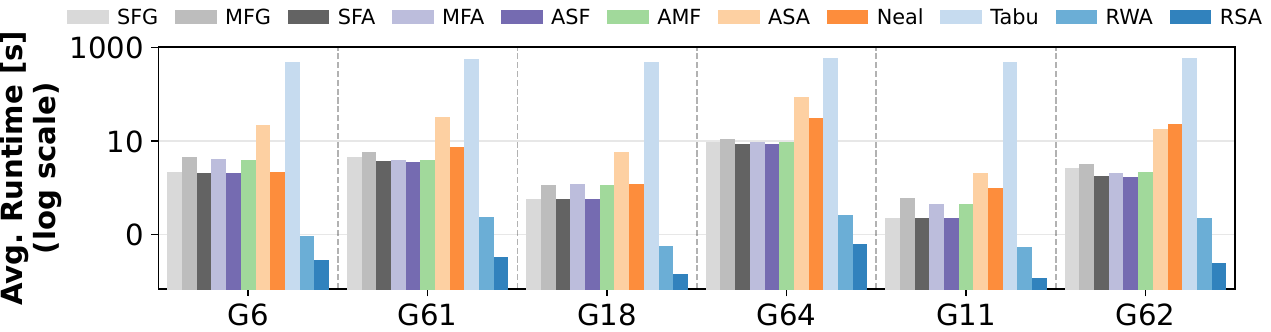}
  \vspace{-2.0em}
  \caption{Runtime of each algorithm corresponding to Table~\ref{tab:gsetmcp}.}
  \label{fig:runtime}
  \vspace{-0.5em}
\end{figure} 

Table~\ref{tab:gsetmcp} shows that Snowball's RWA and RSA achieve the highest solution quality, and Figure~\ref{fig:runtime} shows that their runtime is also the fastest, indicating rapid convergence.

\subsubsection{Time-to-Solution on $K_{2000}$ Max-Cut}
Time-to-solution (TTS) is commonly used to quantify how long a stochastic solver must run to reach a target success probability $p$. Each run is modeled as a Bernoulli trial that finds a (near-)optimal solution with probability $P_a(t_a)$ within computing time $t_a$. Assuming independent and identically distributed runs, the probability of observing at least one success in $R$ runs is $P_{\ge 1}(R) = 1 - (1 - P_a(t_a))^R$. Enforcing $P_{\ge 1}(R) \ge p$ and solving for the smallest $R$ yields $R \ge \frac{\ln(1 - p)}{\ln(1 - P_a(t_a))}$~\cite{boixo2014evidence, ronnow2014defining}.
The time-to-solution is then defined as
\begin{equation}
\mathrm{TTS}(p)
= t_a \,\frac{\ln(1 - p)}{\ln\!\bigl(1 - P_a(t_a)\bigr)}.
\end{equation} For a fair comparison with prior work, the cut value threshold for $K_{2000}$ is set to 33{,}000 when computing TTS(0.99), following prior state-of-the-art results~\cite{10609617,9062965,doi:10.1126/science.aah4243,goto2019combinatorial}.

\setcounter{figure}{14}
\begin{figure*}[!b]
  \centering
  \vspace{-1.5em}
  \includegraphics[width=0.9\textwidth]{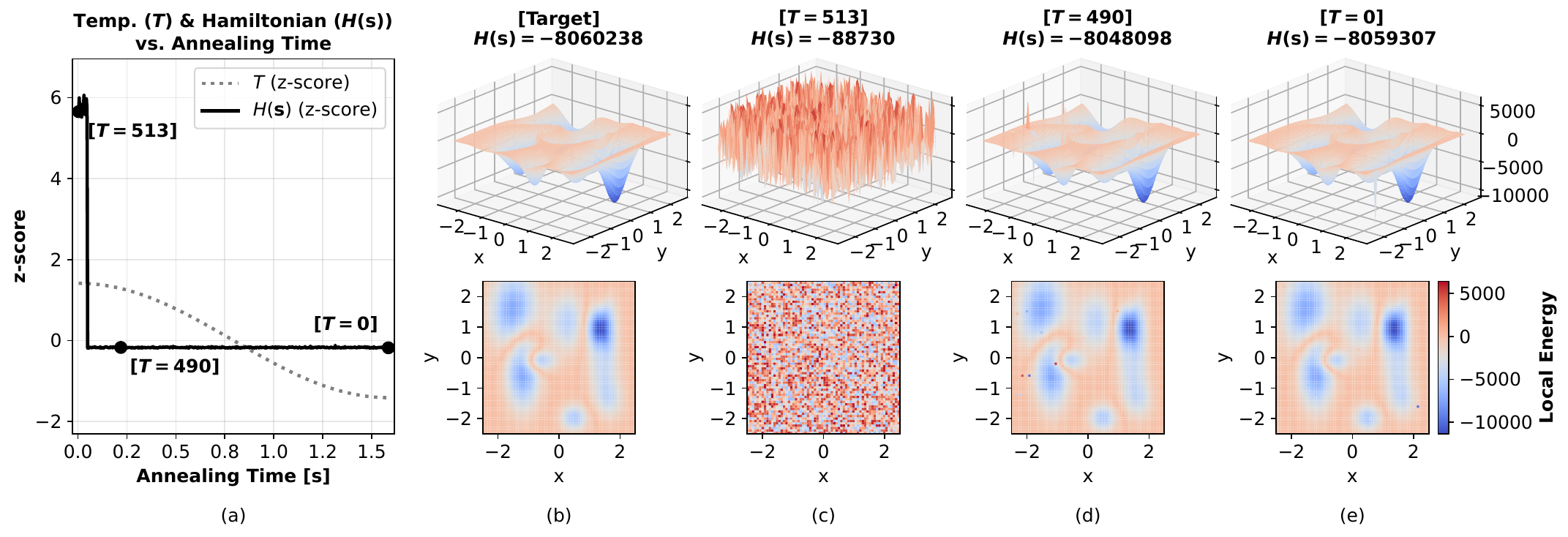}
  \vspace{-1.0em}
  \caption{Hamiltonian in 2D and its 3D surface. (a) Cosine annealing schedule. (b) Target landscape encoded to 16-bit. (c) High \(T\): near-random landscape. (d) Cooling: landscape aligns with target. (e) Low \(T\): recovered landscape almost matches the target (\(99.5\%\) bit-plane accuracy).}
  \label{fig:mot}
\end{figure*}

\begin{table*}[t]
\centering
\caption{Comparison of TTS(0.99) on the $K_{2000}$ Max-Cut Instance}
\label{tab:comp}
\resizebox{\textwidth}{!}{%
\begin{tabular}{|l|cc|c|c|cc|cc|cc|}
\hline
Ising Machine &
  \multicolumn{2}{c|}{Neal~\cite{dwave_neal}} &
  CIM~\cite{doi:10.1126/science.aah4243} &
  SB~\cite{goto2019combinatorial} &
  \multicolumn{2}{c|}{STATICA$^{\dagger}$~\cite{9062965}} &
  \multicolumn{2}{c|}{ReAIM$^{\ddagger}$~\cite{10609617}} &
  \multicolumn{2}{c|}{\textbf{Snowball}} \\ \hline
Hardware Type &
  \multicolumn{2}{c|}{CPU} &
  Optics &
  FPGA &
  \multicolumn{2}{c|}{CMOS} &
  \multicolumn{2}{c|}{CMOS} &
  \multicolumn{2}{c|}{\textbf{FPGA}} \\ \hline
Computing Time $t_a$ [ms] &
  \multicolumn{1}{c|}{4610} &
  5646 &
  5 &
  0.5 &
  \multicolumn{1}{c|}{0.13} &
  0.48 &
  \multicolumn{1}{c|}{0.15} &
  0.23 &
  \multicolumn{1}{c|}{\textbf{0.128}} &
  \textbf{0.085} \\ \hline
Probability $P_a(t_a)$ &
  \multicolumn{1}{c|}{0.38} &
  0.77 &
  0.02 &
  0.04 &
  \multicolumn{1}{c|}{0.07} &
  0.77 &
  \multicolumn{1}{c|}{0.47} &
  0.8 &
  \multicolumn{1}{c|}{\textbf{0.99}} &
  \textbf{0.99} \\ \hline
TTS(0.99) [ms] &
  \multicolumn{1}{c|}{44413} &
  17693 &
  1139.74 &
  56.14 &
  \multicolumn{1}{c|}{8.23} &
  1.5 &
  \multicolumn{1}{c|}{1.11} &
  0.68 &
  \multicolumn{1}{c|}{\textbf{0.128}} &
  \textbf{0.085} \\ \hline
\end{tabular}%
}
\\[0.5ex]
\raggedright\footnotesize
$^{\dagger}$~Since the STATICA hardware supports up to 512 spins, the reported values are obtained from the 2K-spin STATICA simulator~\cite{9062965}.\\
$^{\ddagger}$~ReAIM synthesizes its architecture but does not fabricate a prototype; therefore, the reported values are obtained from its simulator~\cite{10609617}.
\vspace{-1em}
\end{table*}

Table~\ref{tab:comp} summarizes the TTS(0.99) values for the $K_{2000}$ Max-Cut instance reported in previous works, together with the measured data for Snowball. Snowball operates at a kernel frequency of 300\,MHz, and the number of steps in this experiment is set to 100, following prior work~\cite{9062965}. The left Snowball column reports the TTS(0.99) value for the parallel mode, and the right column reports the TTS(0.99) value for the sequential mode. As discussed in Section~\ref{subsec:topo}, thanks to the all-to-all topology, a single-spin update can influence multiple neighbors, so Snowball’s single-spin update does not degrade the TTS for either random-scan spin selection or roulette-wheel spin selection. Since these modes share the same datapath, as described in Section~\ref{subsubsec:hw}, their TTS values are similar. Figure~\ref{fig:speedup} shows that, for the $K_{2000}$ Max-Cut instance, Snowball’s sequential mode achieves a TTS(0.99) speedup of 208{,}153$\times$ over Neal~\cite{dwave_neal} and 8$\times$ over the state-of-the-art ReAIM~\cite{10609617}.

\setcounter{figure}{12}
\begin{figure}[h]
  \centering
  \vspace{-1em}
  \includegraphics[width=\columnwidth]{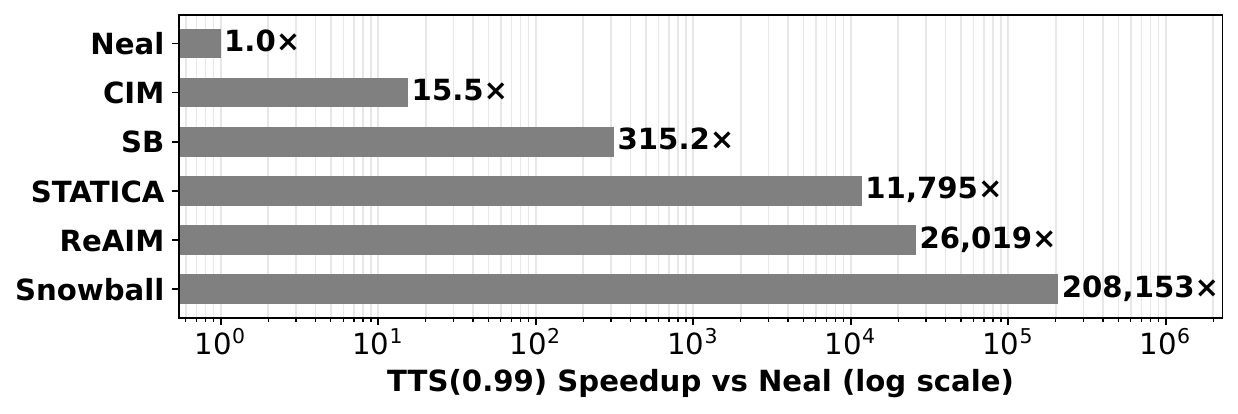}
  \vspace{-2.0em}
  \caption{Speedup in TTS(0.99) achieved by state-of-the-art Ising machines relative to the Neal baseline~\cite{dwave_neal} on a $K_{2000}$ Max-Cut instance.}
  \label{fig:speedup}
  \vspace{-0.5em}
\end{figure} 

Figure~\ref{fig:bound} shows that the kernel-only runtime (excluding DMA) and the end-to-end runtime (including DMA) nearly perfectly overlap, indicating that Snowball is compute-bound rather than memory-bound. Furthermore, comparing the end-to-end runtime with the naive approach (which applies no incremental updates) shows that the incremental-update scheme effectively mitigates memory-bandwidth overhead.

\setcounter{figure}{13}
\begin{figure}[h]
  \centering
  \vspace{-1em}
  \includegraphics[width=\columnwidth]{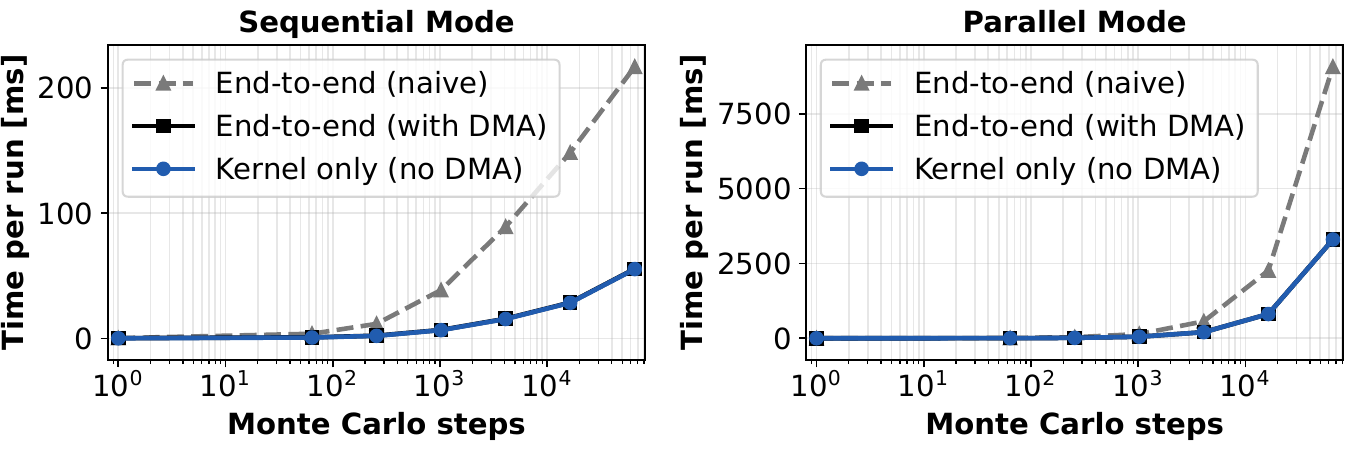}
  \vspace{-2.0em}
  \caption{Kernel-only (excluding DMA) vs.\ end-to-end (including DMA) runtimes across Monte Carlo steps; their overlap indicates that the implementation is compute-bound. The ``Naive'' baseline applies no incremental updates.}
  \label{fig:bound}
  \vspace{-0.5em}
\end{figure}

Figure~\ref{fig:mot} shows that, thanks to bit-plane decomposition (Section~\ref{subsubsec:bitp}), Snowball scales with bit-width and reconstructs a \(64\times64\) field at 16-bit precision (visualized as a 3D surface), achieving \(99.5\%\) pixel-wise agreement in our runs (bit-plane accuracy; exact 16-bit pixel matches between recovered and target).

%% file: sections/conclusion.tex
\section{Conclusion}
Building on an in-depth systematic analysis of key Ising-machine design considerations, this work introduces Snowball, a digital, scalable, all-to-all coupled Ising machine that integrates dual-mode Markov chain Monte Carlo spin selection and asynchronous spin updates, implemented on an AMD Alveo U250 accelerator card. Experimental results demonstrate that Snowball achieves an 8$\times$ reduction in time-to-solution compared to state-of-the-art Ising machines.